# A Joint-Reasoning based Disease Q&A System

PRAKASH CHANDRA SUKHWAL, Department of Information Systems and Analytics, National University of Singapore
VAIBHAV RAJAN, Department of Information Systems and Analytics, National University of Singapore
ATREYI KANKANHALLI, Department of Information Systems and Analytics, National University of Singapore

Medical question answer (QA) assistants respond to lay users' health-related queries by synthesizing information from multiple sources using natural language processing and related techniques. They can serve as vital tools to alleviate issues of misinformation, information overload, and complexity of medical language, thus addressing lay users' information needs while reducing the burden on healthcare professionals. QA systems, the engines of such assistants, have typically used either language models (LMs) or knowledge graphs (KG), though the approaches could be complementary. LM-based QA systems excel at understanding complex questions and providing well-formed answers, but are prone to factual mistakes. KG-based QA systems, which represent facts well, are mostly limited to answering short-answer questions with pre-created templates. While a few studies have jointly used LM and KG approaches for text-based QA, this was done to answer multiple-choice questions. Extant QA systems also have limitations in terms of automation and performance. We address these challenges by designing a novel, automated disease QA system which effectively utilizes both LM and KG techniques through a joint-reasoning approach to answer disease-related questions appropriate for lay users. Our evaluation of the system using a range of quality metrics demonstrates its efficacy over benchmark systems, including the popular ChatGPT.

## 1 INTRODUCTION

The Internet has become a leading source of health information, with around 4-5% of global searches being related to healthcare [45]. In the United States, nearly 70% of the population goes online to search for medical information, for instance to find answers to their questions about diseases, symptoms, and remedies [69]. These sources, including online forums and websites, are popular amongst users due to their ease of access [52, 73]. However, there are several challenges in using them.

First is the issue of *misinformation*, as there is often contradictory or misleading information online

[69]. Much of the health-related content online is user-generated and could potentially misguide users due to inaccuracies [34], failure to update information, and lack of medical experts to validate the advice [36]. Second is the problem of *information overload*, since typically too much information is available in response to online queries, leading to confusion and stress for users [25]. Third is the issue of *low readability* of medical content for lay readers [67], where readability refers to the attribute of being concise and comprehensible to lay users [25]. While scholarly medical articles available online can be accurate and up-to-date, they contain technical jargon and complexity that is inappropriate for lay users [41].

The above issues highlight the need for *automated medical question-answer (QA) assistants*. These are automated software that use natural language processing (NLP) and related techniques to synthesize relevant information from online sources, often by building a structured knowledge base (KB) [81]. The information needs to be collected and updated automatically from reliable knowledge sources to be presented via readable answers to users' health-related questions. Other than lay users, medical professionals can also benefit from such automated QA assistants. With increasing patient queries, it is burdensome for doctors to address them [73]. A QA assistant can offer easy-to-access information to patients [34] and alleviate the workload of medical professionals.

A key component of such QA assistants are the core QA systems, which receive user questions and provide answers to the dialog interface. While considerable progress has been made in QA techniques, these systems face limitations in their automation and performance [31]. The reasons include poor quality or incompleteness of information in the QA system's KB, inability of the system to select a quality answer [34, 50], and high cost of domain customization [40]. We address these challenges by jointly leveraging advances in *knowledge graphs* and *language model* techniques.

Knowledge graphs (KGs) represent concepts and their relationships through a graphical structure of nodes (or entities) and edges (or relations) [32]. An edge and its two connecting nodes form a triplet representing a "fact", e.g., "*Diabetes* is a *disease*" has two nodes (in italics) connected by the "is a" relation. KGs are useful for knowledge representation and retrieval tasks. In the medical domain, they have been used for various applications such as early disease detection [54], predicting adverse effects of drugs [18], as well as biomedical QA for healthcare professionals [24]. In contrast to KGs, language models (LMs) are probability-based models that learn statistical properties of sequential distribution of words in text documents [6]. By learning from large text corpora (collections) and encoding knowledge in a manner different from KGs, LMs are useful for natural question answering, as seen in the recent popularity of ChatGPT[1].

---

[1] https://openai.com/blog/chatgpt/

Both KG and LM-based approaches have their merits and drawbacks. LMs excel at capturing context in more complex natural language queries and in answer simplification by paraphrasing [77]. However, they need large amounts of data for fine-tuning, provide multiple potential answers from which it is difficult to choose the final answer [76], and are prone to hallucinations i.e., factual inconsistencies [33]. KGs perform well in answering short questions as they are able to represent facts. However, this is not the case for longer natural language questions, as they often rely on pre-created templates for questions and answers [34]. KG-based QA systems are typically unable to handle ambiguity in user questions, which often stems from incompleteness of the underlying KB [2, 31].

Thus, LM and KG techniques have complementary strengths, due to which researchers have explored joint reasoning approaches that involve both methods for QA. However, prior joint reasoning approaches for text-based QA have been applied to multiple-choice questions, e.g., References [10, 66], which differs from our context. Particularly, lay users' questions about disease do not adhere to fixed templates for questions or answers and instead require a system capable of understanding natural language queries to generate comprehensible and quality answers. As our review later indicates, the design and evaluation of such medical QA systems for lay users remains unexplored. Motivated thus, we address the following research question: *How can KG and LM approaches be combined to design an automated system for accurate and comprehensible medical QA for lay users?*

In this paper, we develop a system that effectively utilizes both KG and LM techniques for medical QA. Designed specifically for lay users, we create a KB from documents extracted from reliable online healthcare websites for the public. The extensible design of our system supports regular updates, with evolving medical knowledge. We address the issue of scarcity of domain-specific data through unsupervised learning for medical QA generation, which, in turn, is used to obtain more accurate sentence representations from the documents. Leveraging advanced deep learning models for information retrieval (IR) and NLP, we retrieve relevant documents for a given user query from our KB and use state-of-the-art LMs adapted for the medical domain to generate candidate answers. We build our own KG using medical documents for lay users, ensuring its completeness through appropriate preprocessing steps. The KG is used in the final joint reasoning algorithm to select the best answer from candidates generated by the LM. Thus, our disease QA system can produce accurate answers in natural language, free from hallucinations, and without the use of pre-determined question or answer templates.

We comprehensively evaluate our system for multiple chronic disease-related questions. Specifically, its performance is compared against three carefully chosen medical QA baselines for lay users: (1) a system for short QA [25], (2) a system for long QA [22], and (3) the popular ChatGPT system capable of handling both short and long QA. At the time of writing, ChatGPT (version 3.5) was available as a beta release and used for comparison. We carried out a rigorous evaluation of our system's answers using a range of metrics covering *lexical similarity, semantic similarity, readability*, and *contradictions*. On the popular lexical metric ROUGE [46], our system outperformed the three baselines (median scores 0.13, 0.17, and 0.19 respectively) with a median score of 0.36. On the semantic metric BERTScore [78], our system outperformed the baselines (median scores 0.84, 0.86, and 0.86 respectively) with a median score of 0.89. Compared to ChatGPT, our answers scored higher (55.47 vs 68.6) on the Flesch-Kincaid readability score [38]. Additionally, we performed multiple ablation studies as evidence of our design choices. Overall, the evaluations clearly demonstrate the merits of our system that integrates KG and LM approaches to provide easy-to-comprehend and contradiction-free, quality answers to lay users.

To summarize, we contribute by developing a novel disease QA system which: (1) effectively utilizes both LMs and KGs through a new joint-reasoning approach; (2) offers a fully-automated, template-free, easy-to-update design, leveraging on advanced deep learning models in IR and NLP; (3) can generate accurate, lay-user-friendly answers devoid of jargon. By doing so, it mitigates the limitations of LM-based medical QA systems that are prone to hallucinations and broken sentences in their answers. It also mitigates the limitations of KG-based medical QA systems that rely on pre-determined templates for QA and struggle with long natural language queries. We carry out a step-by-step system design, followed by a thorough evaluation over a range of metrics for multiple chronic disease-related questions, which demonstrates the efficacy of our system.

The rest of the paper is organized as follows. In the next section, we discuss related work and the gaps in research. In the following section, we describe our design framework and steps, including the data sources used. Benchmark and ablation results are presented next, followed by a discussion of our study's limitations and suggested future research directions. We conclude with a summary of our contributions.

## 2 RELATED WORK

We conducted a literature review on medical QA systems research from related disciplines (details are in Appendix A). In this section, we discuss the relevant literature and research gaps. An overview of LMs, domain adaptation, and evaluation metrics is provided in Appendix B as background knowledge.

Prior QA systems can be categorized into: (1) KG-based; (2) LM-based; and (3) joint reasoning systems that use both KG and LM techniques, which we review in that order. KG-based QA systems have been designed for both patients and medical professionals. Systems for patients/ the public have provided short answers about diseases and symptoms [31], answered specific procedural questions on treatment [1], and provided food suggestions [15]. For medical professionals, QA applications include question formulation for identification of high-risk patients [2] and answering bio-medical research questions [24].

We find two main research gaps in the KG-based medical QA systems literature. *First*, the problem of KG disconnectedness persists [2, 31], where relevant nodes are not connected, since relations or synonyms are not inferred from knowledge sources. While KGs are relatively easy to update and perform well at finding facts due to their graphical schema, they struggle with understanding long natural language queries or queries using synonymous entities. This is a problem for KG-based systems in general [68]. *Second*, prior work has utilized pre-created question templates from user inputs [1, 15, 24] or pre-created answer templates for answer generation [2, 24, 31]. This allows for handling specific types of questions and short answers, instead of natural QA involving longer questions or answers. *Last,* prior research is lacking in developing disease KGs from reliable data sources devoid of complex medical jargon, which we aim to do.

LM-based systems for natural QA can be: (1) *extractive* [14], where answers are assumed to be present in a given KB and the system extracts the relevant string (span of text) as the answer, or (2) *generative* [43], where the QA system tries to infer and construct an answer based on its understanding of the question and the potential matching sections (or text passages) in a given KB. We use a generative LM approach as it performs better for longer natural language QA [77], while using the KG approach to address LMs' limitations of hallucination [33]. In the healthcare domain, LM-based QA systems have been used to engage patients in natural conversations for promoting health outcomes [79] and to respond to user queries during the covid-19 pandemic through providing short summaries from relevant text databases [22, 25].

We find three key research gaps in the LM-based medical QA systems literature. *First*, there is a lack of domain adaptation of LMs [22, 25], which is needed to obtain domain-specific, quality answers. *Second*, LMs produce multiple candidate answers that need filtering to obtain an accurate and unambiguous answer. The problem of automatically identifying the most relevant (or accurate) answer from the candidate (top-n) answers has not been solved well [21, 73]. Selecting the first candidate answer often results in ambiguous, incorrect or broken answers, including with hallucinations [22, 25]. *Third,* as highlighted before, fully automated QA systems are needed to obtain efficiencies for users and developers. However, prior research has relied on manual, hand-crafted rules for answer selection [22, 25].

Joint reasoning approaches involving both KG and LM techniques have been used in other QA application areas, such as to support video QA [35, 44], where users ask questions about videos in natural language. In text-based QA, such approaches have been restricted to answering multiple-choice questions [10, 66, 77, 80]. The general approach in these studies is to construct a representation for each answer option from an answer-specific substructure of a given KG, e.g., from a subgraph or a path in the KG, and then use LM-based embeddings of the questions jointly with the answer representations to find the most probable answer from the given options. Such methods are not applicable in our case, which lacks answer options.

Additionally, there are research gaps related to the design and evaluation of healthcare QA systems. *First*, our review indicated that prior research typically does not provide a step-by-step system design, including evaluation of each component. *Second*, system evaluations are usually restricted to lexical and semantic metrics. However, readability and contradictions in system responses also need to be assessed [34, 41]. In Table 1, we present a summary of the identified gaps, which this study aims to address.

Table 1: Research Gaps Identified in the Medical QA Systems Literature

| Gap No.* | QA Component | Gap Identified | Implications for QA System |
|---|---|---|---|
| Gap-1 | KG | Limited ability to address disconnectedness of KGs | - Inability to handle ambiguity of terms in user questions |
| Gap-2 | KG | Reliance on template-based QA | - Limited answering ability |
| Gap-3 | LM | Limited domain adaptation for relevant answers | - Limited ability to understand domain specific user questions. |
| Gap-4 | LM | Ambiguous, incorrect, broken answers in system responses including hallucination | - Unclear or incorrect answer generation |
| Gap-5 | LM | Lack of automated analytical models, reliance on manual hand-crafted rules for answer selection | - Unable to select a quality answer from multiple candidate answers generated |
| Gap-6 | System Design and Evaluation | Absence of step-by-step system design and evaluation | - Focus on overall system testing but sub-components may not be assessed |
| Gap-7 | System Evaluation | Mainly lexical or semantic metrics for evaluation | - Other important metrics of answer quality e.g., readability, contradictions, are not evaluated |

*The gaps are numbered for ease of reference in our subsequent discussions*

## 3 SYSTEM DESIGN

### 3.1 Data Collection

Our first source of raw data comes from Mayo Clinic (mayoclininc.org), which is a reliable and lay-user-friendly source of health information [11]. From this source we collected publicly accessible data on several aspects of diseases[2] such as *disease overview, symptoms, causes, risks, treatments, diagnosis*, and *prevention*. We provide more details of this data in Appendix C, Table C1.

Our second data source is the MedQuAD [5] repository on GitHub[3], which was shared publicly by NIH researchers. The data spans over 12 NIH websites (e.g., cancer.gov, niddk.nih.gov, GARD, MedlinePlus Health Topics) and contains 47,000+ medical QA pairs. This source provides a rich and diverse set of questions classified under topics such as *about disease, at risk, cause, diagnosis and test, symptom, treatment,* and *other*. We utilized data related to chronic diseases from this source, since the remaining data is mainly related to rare diseases, which are not the focus of the current study. We provide more details of the MedQuAD data in Appendix C, Table C2. Next, we present our design framework.

---

2 https://www.mayoclinic.org/diseases-conditions
3 https://github.com/abachaa/MedQuAD

## 3.2 Design Framework

Our design follows a multi-step approach to QA system construction. In Step-1, we pre-process the collected data to create three crucial sub-components of our system: (1) a data base of text documents stored in vector format (VDB), (2) a knowledge graph (KG), and (3) a fine-tuned sentence transformer (see Step-1; Figure 1). Specifically, the VDB plays the role of our system's KB with all the text documents stored in a vector (or numeric) format. Also, since the medical field is characterized by an exponential growth of knowledge [64], it is important for us to develop a QA system that facilitates regular updates. Our design allows such updates of the VDB by adding new or updated data sources as the input *Raw Data* in Step-1 as shown in Figure 1. Regular updates from reliable health data resources helps enhance the quality and reliability of our VDB. Steps 2 and 3 are related to generating answers to user questions. In Step-2, a given user question in natural language is used to retrieve relevant documents (also known as *contexts*) from the VDB. Together with the user question, the retrieved documents are passed to a large generative LM which generates 5 candidate answers, which is a typical number of candidates [27] (see Step-2; Figure 1). Last, in Step-3, we implement our joint-reasoning approach for selecting a final answer. Here, using a scoring mechanism, we rank the 5 candidate answers to select a final answer to be shown in response to the given user question (see Step-3; Figure 1). Thus, we generate the final answer without the use of any preset answer template (*Gap-2*). We now elaborate each step in the framework.

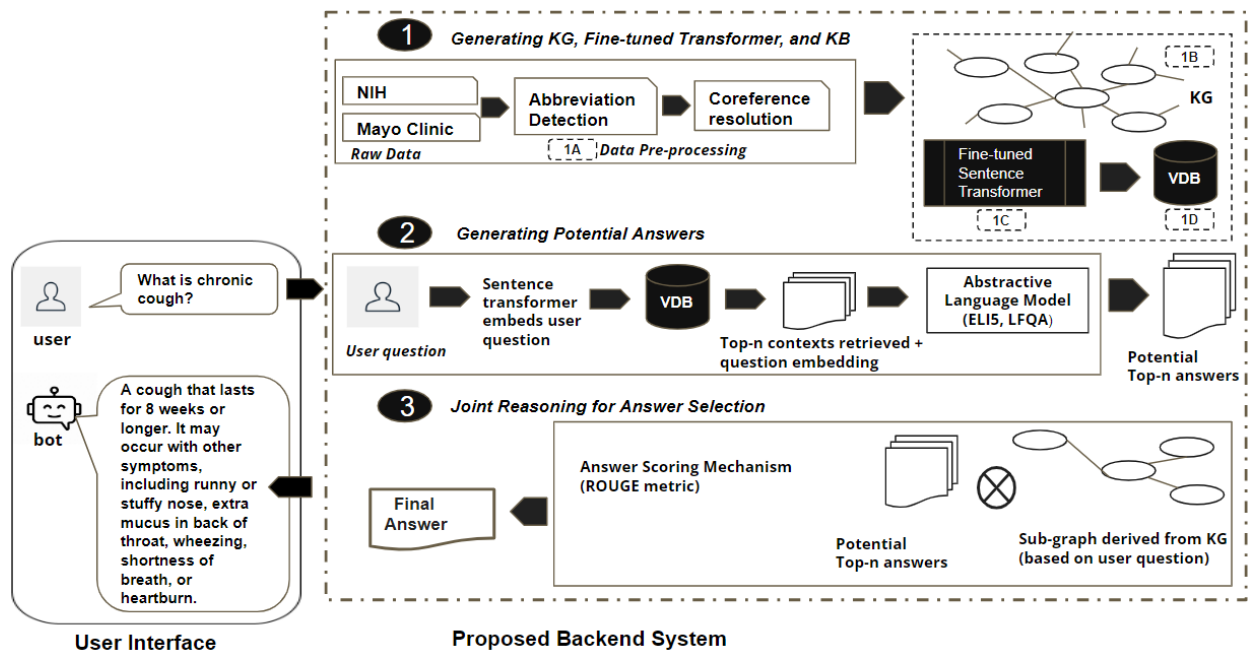

Figure 1: Design Framework for our proposed Question-Answer System

## 3.3 Design Steps

*3.3.1 Data Pre-processing (Step 1A)*

Data gathered from the two sources, Mayo Clinic and MedQuAD, are first passed through two pre-processing steps – abbreviation detection and coreference resolution (see Step-1; Figure 1). Medical data is characterized by inconsistency in the terminologies used in different information sources and by different users [30]. The plethora of abbreviations used in source documents poses a problem for QA systems, as they are often unknown to users and lead to confusion [39]. Abbreviation detection replaces relevant medical abbreviations with their full forms e.g., 'AD' is replaced by 'atopic dermatitis'. We used the Scispacy[4] library (designed for medical text) for abbreviation detection.

Coreference resolution replaces all reference pronouns of a main entity by the name of the main entity to reduce ambiguity from individual sentences e.g., "*Eczema is a prevalent disease in Asia. It is particularly found in infants*" is changed to "*Eczema is a prevalent disease in Asia. Eczema is particularly found in infants*". Coreference resolution helps in linking data pertaining to the same main entity during KG creation (Step 1B) e.g., in the absence of coreference resolution, we will get graph triples ('*Eczema*' is '*a disease*') and ('*It*' is '*found in Asia*'), where the source information of the second triple is lost and not usable in subsequent steps. But with coreference resolution, we get informative triples. This process connects the nodes correctly and addresses the problem of disconnected nodes in a KG (*Gap-1*).

*3.3.2 Generating our KG (Step 1B)*

We chose to construct our own disease KG instead of using pre-existing KGs for multiple reasons. *First*, the publicly available medical KGs such as SemMedDB [37] or Hetnet [29] are general-purpose KGs generated from sources including scientific literature e.g., from the PubMed database. Thus, they contain terms that are difficult for lay users to understand. *Second*, they do not always contain data on all the important features of a disease (e.g., causes, symptoms, risks, preventions, diagnosis etc.). For instance, we observed that SemMedDB contains the feature 'cause', which was useful for us but other features, such as 'treatment', were missing. Similarly, Hetnet only provided information on the feature 'symptom'. In addition, we examined Google health KG [57] but saw similar problems of missing disease-related features such as causes, diagnosis, and risks.

---

4 https://spacy.io/universe/project/scispacy/

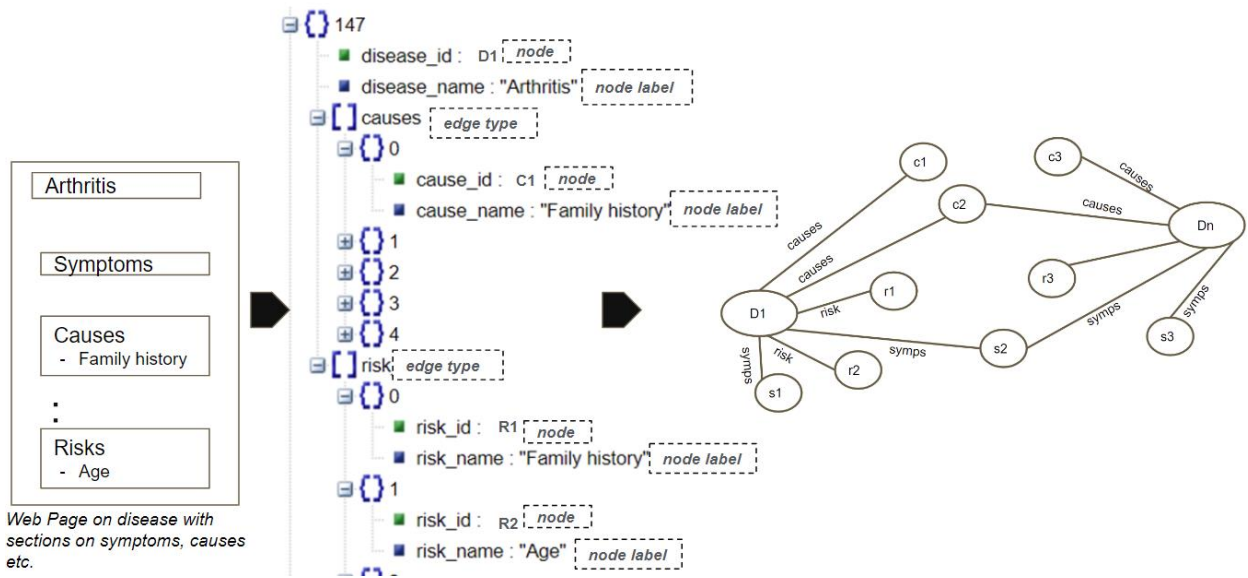

Figure 2: KG Construction from Raw Text

A total of 1,124 diseases are currently contained in our KG which can be expanded with more data in future. We generated our KG using the pre-processed data from Mayo Clinic (see Step-1; Figure 1) pages on specific diseases. Each disease is a node (e.g., D1, D2) in the KG with disease name used as node label (e.g., 'Arthritis'). For each disease, the text content of online pages is collected from sections in the webpage on *symptoms, causes, and diagnosis, treatment*, *prevention*, and *risk factors* (see sample page structure in Figure 2; left). For each section, a node is created for each word/phrase within the section, labeled by the word/phrase itself (see Figure 2; center). The heading of the section (e.g., 'Causes') defines an edge type (or relation) and each of the word nodes within the section are connected to the disease node via an edge of this type e.g., node C1 (label: 'Family history') is connected via the 'causes' edge to node D1 (label: 'Arthritis') (see generic representation in Figure 2; right). Additionally, using Scispacy and information from the unified medical language system (UMLS[5]) database, we obtained a concept ID (CUI) for each disease label in the KG. UMLS provides a set of centralized health and biomedical vocabularies to promote the creation of interoperable systems and services. A node is created for each CUI and edges are added between each disease and its CUI node. Thus, synonymous diseases are connected via CUI nodes in the KG.

*3.3.3 Generating a Fine-tuned Sentence Transformer and Vector DB (Steps 1C and 1D)*

One of the key steps in any natural language QA is to perform question understanding and produce good quality sentence encodings, a mechanism by which a natural language sentence is encoded in vector (or numeric) format of certain dimension (e.g., 384- or 768-dimensional vector per sentence). These vectors can be used to find other semantically similar text documents from the VDB using measures such as cosine similarity between pairs of vectorized texts. The current state-of-the-art methodology for obtaining such encodings is to use neural sentence transformer models [59]. In our proposed system, we used the popular mpnet-base[6] sentence transformer model from Microsoft.

---

[5] https://www.nlm.nih.gov/research/umls/index.html

Fine tuning a sentence transformer on text from a particular domain, in this case medical (e.g., using Mayo Clinic data), enhances the ability of the sentence transformer to better represent pieces of text (numerically) from that domain [26]. The fine-tuning process needs a large number of QA pairs in a specific format. Since we did not have pre-available QA pairs we used an unsupervised technique called GenQ to generate such pairs. GenQ consists of a series of connected steps (see Figure 3) necessary to generate the required QA pairs from text. GenQ takes in paragraphs of text and, internally using a LM, generates QA pairs which are in the required format for the fine-tuning task. Subsequently, these QA pairs can be used to fine-tune any given sentence transformer for the QA task.

To implement GenQ, we used a large Google T5 LM [55] specifically trained to generate QA pairs to fine-tune our sentence transformer model. The pre-processed text data is used for fine tuning. *First*, from all the pre-processed text (after Step 1A), a set of 7,000+ paragraphs are generated and passed to the T5 model. The T5 model in turn generated over 31,000 QA pairs (see Figure 3) which were used to fine-tune the mpnet-base model as part of our domain adaptation (***Gap-3***). More details on the implementation can be found in Appendix D1. Towards the end of Step-1, this fine-tuned sentence transformer is used to convert all the sentences in the input documents collected from Mayo Clinic and MedQuAD resources into vector format and then stored in a VDB (see Figure 1, Step 1D).

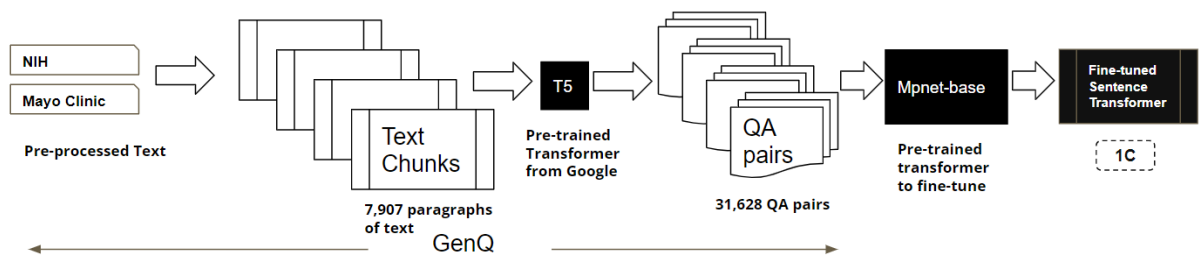

Figure 3: Training the Sentence Transformer LM using GenQ (for query understanding and semantic search)

*3.3.4 Generating Top-5 Potential Answers (Step 2)*

---

[6] https://huggingface.co/microsoft/mpnet-base

For a given user question, we first encode the question using our fine-tuned sentence transformer (see Step-2; Figure 1). The encoded representation of the question is used to search and extract top-5 answer contexts (or answer documents) from the VDB which are most similar to the question using cosine similarity. Later, we pass the encoded question, and the five extracted contexts to large generative LMs for answer generation. We used an average answer length of 120 words similar to earlier studies, e.g., Reference [7] with max length set to 150 words. We evaluated 2 LMs — ELI5 [23] and LFQA for answer generation. These two large LMs based on BART [42] are designed to find or generate an answer from the given answer context and present it in an easy-to-understand language suitable for lay users. Hence, at the end of Step-2 (see Figure 1), for each of the top-5 similar answer contexts we generate a corresponding answer, resulting in top-5 potential answers for further comparisons under joint-reasoning.

*3.3.5 Joint-reasoning for Answer Selection (Step 3)*

To select the final answer, our system performs a series of steps as per our joint-reasoning methodology (see Step-3; Figure 1). *First*, based on the user question, it obtains a relevant subgraph from the overall KG using lexical search. To obtain the subgraph, we first find words in the user question that match either disease names or relations in our KG. Exact or partial string matching is done to match the relations, and a fuzzy matching[7] module is used to match both exact and misspelled disease names. To the identified disease node, we add the remaining incident nodes using the relations matched. In other words, if disease D1 (e.g., 'Arthritis') and relation r (e.g., 'Causes') are matched, then we add all nodes (C1, C2,…,Cn) such that the set of triples (D1,r,C1), (D1,r,C2),…,(D1,r,Cn) exist in our KG (see Figure 2; center, right). This collection of nodes forms our subgraph for relation r. *Second*, all the node labels in the subgraph are collected to form a string, that we call *subgraph-text*. *Last*, a ROUGE-L score is computed between each potential answer (from Step 2) with respect to the subgraph-text. The answer with the maximum F1 score of ROUGE-L is selected as the final answer (see Step-3; Figure 1) and presented to the user.

An illustrative example from our system is shown in Figure 4 for the question – *what are the symptoms of mouth cancer*? The upper block (A) represents the system processing of the user question to obtain the disease name and relevant relation from the user's question. This is followed by fetching the relevant subgraph-text from the KG. The lower left block (B) shows the addition of ROUGE-L scores, with respect to the subgraph-text, to each of the top-5 answers generated in Step-2. Finally, the lower right block (C) shows the final answer based on the maximum score. The text of the answer is formatted before presentation to the end-user.

---

[7] https://pypi.org/project/fuzzywuzzy/

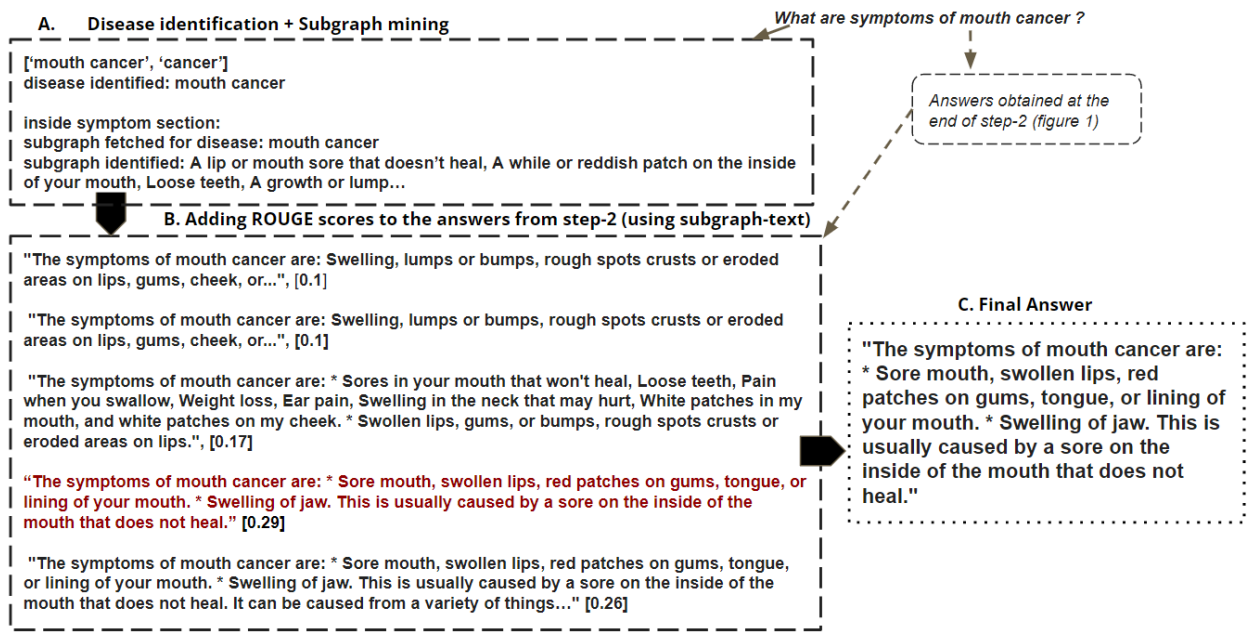

Figure 4: Illustration of Joint-Reasoning Approach in our System

Noteworthy here is that our answer generation and joint-reasoning methodology (*Gap-5*) for obtaining quality answers is different from earlier studies on response ranking [4, 76]. *First*, our choice of data and LM are done specifically for lay users. *Second*, we use our KG as a factual layer to score and select the best possible answer for a user question. For textual QA, joint-reasoning over a KG and a LM thus far has been applied for answering MCQ questions [10, 66, 77, 80]. Other points of difference with respect to earlier medical QA systems [22, 25] are: (1) the use of a domain-adapted sentence transformer models to encode user questions, (2) unlike earlier works, we guard against generation of contradictory information (or hallucination) as well as broken sentences produced by a LM (*Gap-4*) using carefully extracted contexts from our VDB, and a set of hyper parameters from extensive tuning of response generation functions in large LMs. More details on the implementation can be found in Appendix D2.

## 4 RESULTS

We evaluated each design component separately (*Gap-6*) and benchmarked our system-generated answers with respect to three baselines. Our *baseline-1* is the state-of-the-art chatbot system ChatGPT (version 3.5). ChatGPT has been found to facilitate healthcare literacy in recent studies [8, 63]. Our *baseline-2* was constructed to be similar to QA systems described in earlier studies [25] which used extractive QA techniques to answer healthcare questions. Finally, our *baseline-3* was constructed to be similar to a generative healthcare QA system [22]. The quality of final answers from our system and the baselines were compared on multiple metrics such as lexical similarity, semantic similarity, ease of readability, and contradictions (see Appendix B3).

## 4.1 Test Data and Gold Answers

In our experiments, we evaluated our joint-reasoning approach, using common chronic disease-related questions provided by the MedQuAD dataset. For evaluation of answer quality, we selected 10 most common chronic disease[8] related questions. These diseases were - *Alzheimer's disease, Asthma, Chronic Cough, Chronic Kidney Disease, Cancer (Colorectal Cancer, Leukemia), Diabetes, Cardiac issues (Heart Attack, Heart Failure), Obesity, Osteoarthritis,* and *Periodontitis*. In total, we obtained a set of 75 test questions from MedQuAD. Each question was also classified into one of the following seven categories - ***about disease, at risk, cause, diagnosis and test, symptom, treatment***, and ***other*** for more detailed evaluation of system performance for particular categories of questions.

For each question, a standard answer (later referred to as *gold answer*) was generated using arguably the most popular search engine – Google search engine (SE) [71]. With nearly 87 billion monthly visits, Google has the world's largest market share [75]. Amongst all Google searches, 5% of searches globally are for health-related information [3]. Google provides the free SE, which is used not only by the general public but also for a number of research studies, including for identifying patient concerns and investigating health-related phenomena [51, 61]. Moreover, the answers to disease-related questions on Google SE are supported by their own knowledge graph, custom created by experts for quality medical content [57], which makes it an appropriate resource to generate gold answers in our context.

## 4.2 Comparison with Baselines

*4.2.1 Benchmarking on Lexical and Semantic Metrics*

We benchmarked our system responses against the three chosen baselines. *First*, we generated the answers for all 75 questions using the three baseline models. *Next*, we generated our system's responses using joint-reasoning methodology (see sections 3.3.4 and 3.3.5) with the two large LMs in Step 2 (ELI5 and LFQA). *Last*, all the answers from various models were compared with the gold answers (from Google SE) on a lexical metric (ROUGE) and a semantic metric (BERTScore). The results are presented in Tables 2-5 below. Specifically, *model-1* refers to the ELI5 and *model-2* refers to the LFQA based models respectively (in step 2). Eventually we chose the LFQA based model.

On the lexical similarity metric (ROUGE), our system generated answers (for ELI5 and LFQA) scored, on an average, higher than the three baseline models (see Table 2). Also, on the disease category specific questions we observe that our system has all median scores higher than the three baselines (see Table 3) and thus demonstrates superior performance. Particularly, our model-2 scored higher than all other baseline models with statistically significant difference in mean scores.

---

[8] https://www.cdc.gov/chronicdisease/about/costs/index.htm

Table 2: Rouge-L (F1-score) with respect to Gold-answer (overall)

|  |  | **Baseline-1** (ChatGPT) | **Baseline-2** [25] | **Baseline-3** [22] | **ELI5-ans** (joint-reasoning; *model-1*) | **LFQA-ans** (joint-reasoning; *model-2*) |
|---|---|---|---|---|---|---|
| **Gold Answer ROUGE-L F1-score (0-1)** | min | 0.00 | 0.00 | 0.03 | 0.03 | **0.1** |
|  | max | 0.48 | 0.94 | 0.6 | 0.79 | **0.96** |
|  | median | 0.19 | 0.13 | 0.17 | 0.17 | **0.36** |
|  | mean | 0.20 (6.4e-10)[+] | 0.23 (2.0e-05)[+] | 0.18 (1.8e-11)[+] | 0.23 | **0.40** |
|  | std. dev. | 0.09 | 0.26 | 0.10 | 0.18 | 0.24 |

*model-1*: answers obtained by our system using ELI5 model with joint-reasoning;
*model-2*: answers obtained by our system using LFQA model with joint-reasoning – our final model
[+] Values in brackets depict p-value of t-test conducted for all baselines with respect to model-2 (our final model)

Table 3: Rouge-L (F1-score) with respect to Gold-answer (questions split by categories)

|  | **Category** | **Baseline-1** (ChatGPT) | **Baseline-2** [25] | **Baseline-3** [22] | **ELI5-ans** (joint-reasoning; *model-1*) | **LFQA-ans** (joint-reasoning; *model-2*) |
|---|---|---|---|---|---|---|
| **Gold Answer median ROUGE-L F1-score (0-1)** | *about disease* | 0.26 | 0.38 | 0.16 | 0.29 | **0.42** |
|  | *at risk* | 0.23 | 0.14 | 0.12 | 0.18 | **0.47** |
|  | *cause* | 0.24 | 0.13 | 0.17 | 0.24 | **0.32** |
|  | *diagnosis and test* | 0.18 | 0.07 | 0.13 | 0.15 | **0.40** |
|  | *other* | 0.16 | 0.19 | 0.14 | 0.10 | **0.45** |
|  | *symptom* | 0.17 | 0.17 | 0.10 | 0.13 | **0.29** |
|  | *treatment* | 0.15 | 0.00 | 0.12 | 0.10 | **0.17** |

Similarly, on the word-level semantic similarity metric (BERTScore), our system (model-2) responses ranked much higher in both overall mean scores (see Table 4) as well as disease category specific median scores (see Table 5). Here again, model-2 scored higher overall and across all question categories against all other models. Additionally, we also evaluated scores for sentence level semantic similarity using STS metric and found that our system performed better than all the baselines. The detailed STS scores can be found in Appendix D, Tables D2-3.

Table 4: BERTScore with respect to Gold-answer (overall)

|  |  | **Baseline-1** (ChatGPT) | **Baseline-2** [25] | **Baseline-3** [22] | **ELI5-ans** (joint-reasoning; *model-1*) | **LFQA-ans** (joint-reasoning; *model-2*) |
|---|---|---|---|---|---|---|
|  | Min | 0.80 | 0.7 | 0.79 | 0.82 | **0.83** |
|  | Max | 0.90 | 0.97 | 0.91 | 0.94 | **0.99** |
| **Gold Answer BERTScore (0-1)** | Median | 0.86 | 0.84 | 0.85 | 0.86 | **0.89** |
| | Mean | 0.86 (2.6e-13)[+] | 0.84 (3.3e-10)[+] | 0.85 (5.0e-15)[+] | 0.87 | **0.90** |
| | std. dev. | 0.02 | 0.05 | 0.03 | 0.03 | 0.04 |

[+] *Values in brackets depict p-value of t-test conducted for all baselines with respect to model-2 (our final model)*

Table 5: BERTScore with respect to Gold-answer (questions split by categories)

| | **Category** | **Baseline-1** (ChatGPT) | **Baseline-2** [25] | **Baseline-3** [22] | **ELI5-ans** (joint-reasoning; *model-1*) | **LFQA-ans** (joint-reasoning; *model-2*) |
|---|---|---|---|---|---|---|
| **Gold Answer median BERTScore (0-1)** | About disease | 0.88 | 0.88 | 0.88 | 0.90 | **0.93** |
| | At risk | 0.86 | 0.83 | 0.86 | 0.87 | **0.90** |
| | Cause | 0.86 | 0.84 | 0.85 | 0.87 | **0.88** |
| | Diagnosis and test | 0.85 | 0.83 | 0.84 | 0.86 | **0.91** |
| | Other | 0.90 | 0.87 | 0.84 | 0.86 | **0.91** |
| | Symptom | 0.84 | 0.86 | 0.86 | 0.85 | **0.88** |
| | Treatment | 0.80 | 0.78 | 0.85 | 0.84 | **0.88** |

These results, from comparing against multiple baselines, emphasize the superior performance of our proposed joint-reasoning methodology over chosen baselines for disease-related QA. This also demonstrates our system's ability to give clear and concise answers similar to Google SE. On the basis of these results, we chose ***LFQA-ans (joint-reasoning)*** (model-2) as our final model for answer generation component in our proposed QA assistant.

*4.2.2 Benchmarking on Readability and Contradiction*

Other aspects of answer quality evaluated in our work are readability and presence of any contradictory information vis-à-vis the gold answers. Further to the comparisons presented in section 4.2.1, we used a metric called Flesch-Kincaid (FK) score to compute the readability of answers from our chosen model-2 (***Gap-7***). For benchmarking, we computed the Flesch-Kincaid scores for each answer obtained from Google SE, ChatGPT, and our proposed system LFQA-ans (joint-reasoning) model-2. The scores are obtained on a scale of 0.0 to 100 (higher the better reading quality). Table 6 below shows an illustrative example of a question on Chronic Cough asked to both ChatGPT and our system. Our system produced a simple succinct answer for the end user and scored higher on the FK-score metric.

Table 6: Example of a disease question asked to ChatGPT and our system with respective FK-scores

| *Question*: What are symptoms of Chronic Cough? | |
|---|---|
| *ChatGPT*: A cough is a reflex that helps clear the airways of mucus, phlegm, and other irritants. It is a common symptom that can be caused by a variety of conditions, including the common cold, flu, and allergies.<br>Symptoms of a cough may include:<br>Coughing: A cough is a symptom in and of itself. | FK-Score: 73.1 |

| Mucus or phlegm: A cough may be accompanied by the production of mucus or phlegm.<br>Chest discomfort: A cough may cause discomfort in the chest.<br>Shortness of breath: A cough may cause shortness of breath, particularly in people with underlying lung conditions.<br>Hoarseness: A cough may cause hoarseness or a change in the voice.<br>Fatigue: A cough may cause fatigue due to the effort required to cough. | |
|---|---|
| *Our system*: A runny or stuffy nose, a feeling of liquid running down the back of your throat (postnasal drip). Frequent throat clearing and sore throat, hoarseness, wheezing and shortness of breath. Heartburn or a sour taste in your mouth. In rare cases, coughing up blood. | FK-Score: 75.5 |

In Table 7, we see the readability score distribution's summary statistics from our system, ChatGPT and Google SE. Our proposed system, on average, produced a higher number of answers which are easier to read for low language literacy users. In contrast, ChatGPT responses were more verbose and scored lower on the readability metric when compared with the gold answers.

Table 7: Flesch-Kincaid scores for each answer obtained from, Google SE, ChatGPT, and our system

| | | **Google SE** | **ChatGPT-ans** | **Our System (joint-reasoning; model-2)** |
|---|---|---|---|---|
| **Flesch-Kincaid scores (0-100)** | min | 9.22 | 28.03 | 24.44 |
| | max | 96.99 | 77.37 | 96.69 |
| | median | 62.17 | 55.47 | **68.60** |
| | mean | 58.30 | 55.47 (0.37)[+] | **64.06** (0.02) |
| | std. dev. | 17.74 | 12.96 | 16.18 |

[+] *Values in brackets depict p-value of t-test conducted for ChatGPT and our system with respect to Google SE*

Additionally, the distributional overlap between the systems was inspected using kernel density estimate plots (see Figure 5). The percentage overlap shows that the readability of our system's answers (see Figure 5; right) (**78.9%**) is better than that of Google SE as the distribution, including the mode, is shifted to the right. In comparison, ChatGPT answers achieved a lower overlap (**73.7%**) with those from Google SE (see Figure 5; left).

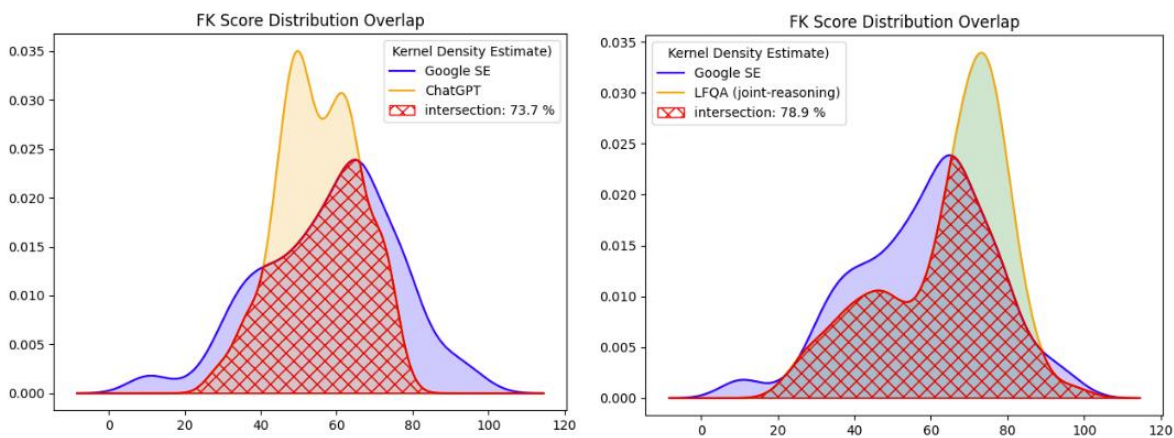

Figure 5: Kernel Density Distribution of gold-ans (blue), ChatGPT (orange), and LFQA-ans (our system in green)

To find any contradictory information in the other model answers with respect to the gold answers, we used the sentence entailment approach (see Appendix E for implementation details). In the entailment scoring, both ChatGPT and our model-2 (joint-reasoning) did not produce any contradictory information with respect to the Google SE gold answers. This shows that our proposed system confined its answers to the given context from the VDB and did not hallucinate or add any contradictory information on its own.

### 4.3 Generating Answers without Joint-reasoning (Ablation Study)

To demonstrate the importance of our joint-reasoning (see step-3; Figure 1) methodology, we performed an ablation study by turning off the joint-reasoning step (i.e., removing step-3; Figure 1) of our system and considered the first among the top-5 answers generated by the large LMs at the end of step-2 (see Figure 1) as the final answer. From the results in Table 8 and Table 10, we can clearly see improvements in answer scores when joint-reasoning methods were used (models 1 and 2). The methodology led to superior performance on both the metrics (ROUGE and BERTScore). Similarly in Table 9 and Table 11 we can see that the median scores improved across all disease categories when joint-reasoning was used. Our model-2 consistently scored higher than all other models with statistically significant difference in mean scores.

Table 8: Rouge-L (F1-score) with respect to Gold-answer (overall)

|  |  | ELI5-ans (No joint-reasoning) | LFQA-ans (No joint-reasoning) | ELI5-ans (joint-reasoning; model-1) | LFQA-ans (joint-reasoning; model-2) |
|---|---|---|---|---|---|
| **Gold Answer ROUGE-L F1-score (0-1)** | Min | 0.03 | 0.03 | 0.03 | **0.1** |
|  | max | 0.70 | 0.6 | 0.79 | **0.96** |
|  | median | 0.12 | 0.17 | 0.17 | **0.36** |
|  | mean | 0.21 (5.5e-09)[+] | 0.18 (1.8e-11)[+] | 0.23 | **0.40** |
|  | std. dev. | 0.12 | 0.10 | 0.18 | 0.24 |

***No joint-reasoning:*** *answers obtained without a joint-reasoning approach;* ***model-1****: answers obtained by our system using ELI5 model with joint-reasoning;* ***model-2****: answers obtained by our system using LFQA model with joint-reasoning*
*+ Values in brackets depict p-value of t-test conducted for models without joint-reasoning with respect to model-2 (our final model)*

Table 9: Rouge-L (F1-score) with respect to Gold-answer (questions split by categories)

|  | Category | ELI5-ans (No joint-reasoning) | LFQA-ans (No joint-reasoning) | ELI5-ans (joint-reasoning; model-1) | LFQA-ans (joint-reasoning; model-2) |
|---|---|---|---|---|---|
| **Gold Answer median ROUGE-L F1-score (0-1)** | *about disease* | 0.13 | 0.16 | 0.29 | **0.42** |
|  | *at risk* | 0.16 | 0.12 | 0.18 | **0.47** |
|  | *cause* | 0.14 | 0.17 | 0.24 | **0.32** |
|  | *diagnosis and test* | 0.17 | 0.13 | 0.15 | **0.40** |
|  | *other* | 0.13 | 0.14 | 0.10 | **0.45** |
|  | *symptom* | 0.11 | 0.10 | 0.13 | **0.29** |
|  | *treatment* | 0.08 | 0.12 | 0.10 | **0.17** |

Table 10: BERTScore with respect to Gold-answer (overall)

|  |  | ELI5-ans (No joint-reasoning) | LFQA-ans (No joint-reasoning) | ELI5-ans (joint-reasoning; model-1) | LFQA-ans (joint-reasoning; model-2) |
|---|---|---|---|---|---|
| **Gold Answer BERTScore (0-1)** | min | 0.8 | 0.79 | 0.82 | **0.83** |
| | max | 0.90 | 0.91 | 0.94 | **0.99** |
| | median | 0.85 | 0.85 | 0.86 | **0.89** |
| | mean | 0.86 (5.0e-15)+ | 0.85 (3.8e-10)+ | 0.87 | **0.90** |
| | std. dev. | 0.02 | 0.03 | 0.03 | 0.04 |

*No joint-reasoning: answers obtained without a joint-reasoning approach; **model-1**: answers obtained by our system using ELI5 model with joint-reasoning; **model-2**: answers obtained by our system using LFQA model with joint-reasoning*
+ Values in brackets depict p-value of t-test conducted for models without joint-reasoning with respect to model-2 (our final model)

Table 11: BERTScore with respect to Gold-answer (questions split by categories)

|  | Category | Baseline-1 (ChatGPT) | ELI5-ans (No joint-reasoning) | LFQA-ans (No joint-reasoning) | ELI5-ans (joint-reasoning; model-1) | LFQA-ans (joint-reasoning; model-2) |
|---|---|---|---|---|---|---|
| **Gold Answer median BERTScore (0-1)** | *about disease* | 0.88 | 0.88 | 0.88 | 0.90 | **0.93** |
| | *at risk* | 0.86 | 0.85 | 0.86 | 0.87 | **0.90** |
| | *cause* | 0.86 | 0.87 | 0.85 | 0.87 | **0.88** |
| | *diagnosis and test* | 0.85 | 0.85 | 0.84 | 0.86 | **0.91** |
| | *other* | 0.90 | 0.84 | 0.84 | 0.86 | **0.91** |
| | *symptom* | 0.84 | 0.85 | 0.86 | 0.85 | **0.88** |
| | *treatment* | 0.80 | 0.84 | 0.85 | 0.84 | **0.88** |

## 4.4 Additional Experiments

We performed additional experiments to evaluate our design choices as well as the impact of components within our system on its performance. Table 12 summarizes the experiments and results. More details can be found in Appendix F.

Table 12: Additional Experiments

| Experiment | Summary | Results |
|---|---|---|
| F1. Evaluation of KG responses for QA | We evaluated our KG (section 3.3.2) as a stand-alone QA resource to examine its ability to answer natural language questions. | The KG is better suited to be used as a collection of facts to refine answers rather than a standalone answer generation resource |
| F2. Evaluating quality of fine-tuned sentence transformer model | To assess the quality of the encodings produced by our newly domain adapted sentence transformer model (section 3.3.3), we compared it with top-3 pre-trained sentence transformers available | Our fine-tuned sentence transformer for medical text achieved a good similarity score of 0.83 in producing sentence encodings comparable to other state-of-the-art sentence transformers fine-tuned on non-medical texts. |

|  | on the sentence transformer website [58]. | Hence, we use our own sentence transformer model for generating sentence encodings and question representations. |
|---|---|---|
| F3. Length-controlled comparison of answer quality | ChatGPT tends to produce verbose answers (words > 250) which can impact metrics such as ROUGE, STS, BERTScore. Hence, we did a length-controlled comparison of answer quality with the ChatGPT system. | In the length-controlled setup, all of our system's mean scores were found to be higher and statistically significant ($p < 0.05$) when compared to respective ChatGPT scores. |
| F4. Generating answers without a VDB (Ablation) | Our VDB (see step-1; Figure 1) is an important part of the proposed system. To highlight its importance, we generated answers to all the questions without using data from VDB to investigate the quality of answers generated. | Results showed that in the absence of contexts from a VDB, the large LMs run the risk of providing confusing, or incorrect answers. This underpins the importance of using a good quality VDB. |

## 5   LIMITATIONS AND FUTURE WORK

There are a few limitations of this work. *First*, since some of our system data is proprietary to Mayo Clinic, we are unable share this dataset publicly. However, our main purpose of using the data was to demonstrate our system design and performance. In future, such data can be collected automatically once requisite permissions are obtained from data owners. *Second*, since all LMs utilize knowledge from the past, they are unable to automatically keep up with the latest advances in human knowledge e.g., the knowledge base of ChatGPT was till 2021 [12]. Going forward, web-crawling from reliable online resources can be integrated within such a QA system to add disease-related information that is missing from the VDB. *Last,* to further assess the quality of answers, system evaluations can be carried out with both lay users and medical professionals. For lay users, perceptions such as relevance and comprehensibility of answers could be measured. For medical practitioners, perceptions such as completeness, informativeness, and truthfulness of answers, could be assessed.

There are several other avenues for future research based on this work. *First*, the current design represents the core engine of a QA system. In order to present a QA system to users, dialog management and user interface components can be designed and added in future, and hosted online to make it more accessible to users. *Second*, such a system design can be expanded to gather user data to personalize the answers according to the user's profile or chat history, with the necessary privacy provisions. *Finally*, there is potential to extend this QA system design to other knowledge domains, such as finance or law, though appropriate KBs would be required for those domains.

# 6 DISCUSSION AND CONTRIBUTIONS

In this study, we presented the detailed steps for designing and evaluating a novel disease QA system. We demonstrated how to build a KG and a VDB from reliable knowledge resources in an automated way. Also, we showed how our novel joint-reasoning methodology can be applied to mine quality answers from a set of top-n potential answers generated by large LMs.

Through this work, we offer research contributions on designing medical QA systems for lay users in the following ways. *First*, we provide a novel joint-reasoning methodology that effectively utilizes a KG and a large LM to provide quality answers for medical questions. To our knowledge, such an approach has not been used in prior text QA studies, which have used either KGs or LMs on their own. *Second*, we present the detailed steps and choices for designing a disease KG with attention to uniqueness and connectedness of nodes. Our system design facilitates periodic update if new data on a disease is added to the online resources, or an additional resource is added altogether (e.g., data from WHO). *Last*, we perform step-by-step system design and thorough evaluation (for each component – KG, sentence transformer, and the overall joint reasoning approach) over a range of quality metrics for multiple chronic disease-related questions. Collectively, these add to IS research on the design of QA systems with the capability of generating correct, readable, template free, and appropriate answers to users' questions.

We also offer a number of *practical contributions* for medical IS on the design of QA systems. *First*, we explicate the data procurement, cleaning, and enrichment process for a disease QA system with lay users in mind. We utilize easy-to-understand, reliable resources like Mayo Clinic and MedQuAD data to build our VDB and KG. *Second*, we demonstrate a complete pipeline on how to fine-tune a custom sentence transformer model from raw text for a disease QA system without the need for labelled data. *Last*, through thorough evaluations using multiple metrics, we show that our system provides quality answers which were ranked higher than those from multiple baselines including the popular ChatGPT. We also supported our system-specific design choices with a set of ablation studies. Our data and design choices have direct practical implications for enabling accurate answers, providing scalable system design, helping patients with information overload, and saving valuable healthcare resources and consultation time by providing 24x7 availability of automated QA of disease-related user questions.

# APPENDICES

## A LITERATURE REVIEW

### A1 Review Methodology

For a comprehensive review, we looked at studies from year 2000 onwards on the Web of Science repository. We adopted a multi-stage process to examine and select relevant papers. *First*, for the above-mentioned outlets, we performed a broad search on both key phrases and abstracts using our search query (see Table A1). *Second*, the abstracts of all the papers filtered from first stage were examined including the keywords to exclude all irrelevant papers. A total of 43 studies were retained after stage two. *Third*, we screened the papers based on the full text to retain relevant papers. In the *fourth* stage, we performed a backward search to include papers relevant to our current study. As a result, a final set of 8 papers specific to healthcare QA were identified for our review.

Table A1: Search and Selection Process

| Search Query | (((("QA", OR 'question answer') AND (Conversational OR Interactive OR Virtual OR Digital OR AI OR 'Artificial Intelligence') AND (Agent OR Assistant)) OR Chatbot) AND ('knowledge graph' OR 'language model' OR 'KG' OR 'LM' OR 'Information Retrieval')) |
|---|---|
| Initial Results (stage 1) | 323 Papers |
| Summary of Exclusion Criteria | • Papers are not from well-ranked (SJR) IS/CS/Medical outlets<br>• The paper mainly focused on other agent systems e.g., simulation or software agents, not relevant to current study<br>• Papers did not focus on question-answering (QA) systems but on conversation capability of chatbots<br>• Papers did not focus on medical domain |
| Papers after abstract filtering (stage 2) | 43 |
| Papers after full-paper review (stage 3) | 7 |
| Final set of papers including papers added via backward search (stage 4) | 8 Papers (5 KG, 3 LM) relevant to medical QA |

## B CONCEPTUAL BACKGROUND

### B1 Language Models

Using large volumes of textual data, LMs model the probability distribution of word sequences, along with the context of surrounding words. Popular LMs such as BERT [20] from Google, RoBERTa [47] from Facebook, and ELECTRA [17] showed near human performance on many downstream NLP tasks, such as identifying named entities from text (NER), sentence completion, emotion classification, and QA. Petroni et al. [53] note that there are several benefits of using LMs as KBs for QA systems – they obviate the need for schema (cognitive frameworks), enable open-domain questions (while KG are usually domain restricted), and are easy to retrain with more data. Also, once created, LMs can acquire additional knowledge from other domains using a technique known as domain-adaptation (or fine-tuning). This helps a LM to improve its performance in domain-specific QA tasks [26]. But LMs also need large amounts of data for fine-tuning, and they are prone to hallucinations i.e., factual inconsistencies [33].

There are two main forms of providing answers to a user question in a natural QA setting- (1) *extractive* [17], and (2) *generative* QA [43]. Extractive QA assumes that given a question, the answer is available in one of the many documents in-line in the KB. The QA system, in this case, tries to make accurate extraction of the answer string (or span of text) and presents it as an answer. In the NLP domain, extractive QA falls under information retrieval (IR) and reading comprehension tasks. On the other hand, generative QA deals with IR and the text generation task and makes no assumption about the answer being directly present in a given KB. The QA system tries to appropriately infer and construct an answer based on its understanding of the question and the potential matching sections (or text passages) of a given KB, which is also the focus of our QA system. The ability to understand a given question in a particular domain is gained via the process of domain adaptation. We discuss this concept in the next section (see B2). In Table B1, we have provided a set of illustrations to better understand extractive versus generative answers for medical questions.

Table B1: Illustration of Extractive versus Generative Answers

*(extractive answers generated using a T5 model; Generative answers using BART-based LFQA model)*

| Question | Extractive Answer | Generative Answer |
|---|---|---|
| What is a treatment for Chronic Cough? | no-answer | Cough is caused by inflammation of airways. There are a number of medications that can be used to treat chronic cough. The most common are: Antihistamines, corticosteroids and decongestants. These drugs are standard treatment for allergies and postnasal drip. |
| How to prevent Overweight and Obesity? | Choosing healthier foods whole grains, fruits and vegetables, healthy fats and protein sources and beverages | There are a few ways to prevent weight gain. The first is to eat a balanced diet. This means eating a variety of foods, including fruits, vegetables, lean meats, and whole grains. If you are overweight or obese, you need to be able to get enough of these foods to meet your nutritional needs. You can do this by eating healthy foods and exercising regularly. Another way to avoid being overweight and obese is by being active. Being active is important because it helps you lose weight. For example, if you exercise regularly, your body will burn more calories than it needs to. |

**B2 Domain Adaptation for LMs**

Pre-trained LMs can be made more domain-specific through domain adaptation (or fine tuning) [26] on text from a particular domain, such as medical in our case e.g., using reliable disease information from Mayo Clinic [11]. This allows an LM to gain a better understanding of the target domain. LMs are neural network models and the process of domain adaptation involves, using additional/new domain specific data, training the top neural network layers of a pre-trained LM model so that the representations learn and adjust using the task-relevant unlabeled data, such as the popular Stanford Question Answering Dataset (SQuAD) [56] released to train LMs for QA.

When trained on large amounts of labeled data (supervised training), the sentence transformer LMs [59] can produce better vector representations of question sentences and answer passages which facilitate the downstream task of answer generation. However, large amounts of labelled data for a particular domain and task are not always readily available or the domain adaptation is cost intensive, which makes supervised training difficult [40]. To address this problem, recent works in NLP propose unsupervised methods for training sentence transformers such as Transformer-based Sequential Denoising Auto-Encoder (TSDAE) [74], and GenQ [70]. Later, in the System Design and Implementation section, we show how we used unsupervised methods to domain adapt our own sentence transformer.

**B3 Metrics for Evaluation**

Generative QA helps solve the information overload issue for end users by scanning through a large number of documents to construct a final answer, but it is imperative to evaluate the quality of answers [34]. During evaluation, in practice, similarity of machine generated answers are tested against pre-provided gold answers [60]. Earlier studies have either used n-gram based or semantic similarity metrics to perform evaluations on system responses. In order to be rigorous in answer quality evaluation, we used a range of metrics to evaluate lexical similarity, semantic similarity, readability, and contradictions.

*Lexical* similarity metrics rely on string comparison popularly known as n-gram comparison. *ROUGE* (Recall-Oriented Understudy for Gisting Evaluation) [46] is one of the most popular lexical metrics. ROUGE ranks a given text on the scale of 0 to 1 (lowest to highest) with respect to the text of a gold answer (or reference text) and was shown to outperform other content-based evaluation metrics [62]. Lexical similarity metrics have a major drawback as these metrics rely on word-level overlaps but do not consider the semantic overlap (texts with similar meanings but different words, which could be through synonymous words or alternative phrasing with similar meanings) between a machine generated answer and a gold answer [41, 60]. To address this gap and capture word-level similarity, researchers proposed the metric *BERTScore* [78] that uses cosine similarity (scale of 0 to 1; lowest to highest) to measure the similarity between a word and its synonyms.

As we noted earlier, low readability in the medical content is a problem especially for lay readers [25, 67]. The metric commonly known as *Flesch-Kincaid score* is used to measure the ease of readability of system generated answers. Additionally, machine generated texts are prone to hallucination [33], which means the addition of any incorrect or unintended text in the generated text. Therefore, tests for any contradictory or misleading sentences in the machine generated answers can be done using a *sentence entailment* approach. This approach uses the fact that two sentence vectors will point in the same direction if they convey the same information and will point in opposite directions if they provide contradictory information [48]. *Sentence entailment* comparison between 2 given sentences produces one of three outputs - positive, neutral, or negative entailment (see Table B2) label for a sentence with respect to the gold answer.

Table B2: Examples of Positive and Negative Entailment

| Gold Answer | Asthma is a **disease with no cure** but can be controlled with medicines. | |
|---|---|---|
| Example Answer -1 | Asthma is **a curable disease** and medicines are usually given to control it. | *negative* entailment |
| Example Answer -2 | Asthma can be controlled using certain drugs. | *positive* entailment |
| Example Answer -3 | Asthma can be found in humans of any age. | *neutral* entailment |

## C DATA DESCRIPTION

Table C1: Data Description of Mayo Clinic - Disease and Conditions Dataset

| Variable | Description |
|---|---|
| disease_name | name of the disease as given on the Mayo Clinic website |
| overview | disease description and overview |
| symptoms | comma separated list of symptoms |
| causes | comma separated list of causes |
| risk factors of disease | key risk factors that cause the disease |
| risk due to disease | key risks caused by a disease |
| at risk | who is most at risk |
| treatment | treatments for a disease |
| prevention | guidelines on disease prevention |
| diagnosis | how the disease is diagnosed by doctors |

Table C2: Data Description of MedQuAD - Medical Question-answer Dataset

| Variable | Description |
|---|---|
| q_type | category/type of a question: about disease, at risk, cause, diagnosis and test, symptom, treatment, and other |
| question | disease related specific question. e.g., what is Anemia? |
| answer | answer provided for the respective questions by NIH researchers |

## D ADDITIONAL DETAILS ON EXPERIMENTS

### D1 Details on fine-tuning sentence transformer using GenQ

On our whole pre-processed data, we generated nearly 31,000 question-answer pairs and stored them in batches of flat files. Using these pairs, we fine-tuned an mpnet-base sentence transformer model pre-trained on 1 billion+ training pairs from Microsoft research [65]. We performed the training process using multiple negative ranking loss (MNR) [28] as a cost function. The whole process of training for 5-epochs on an Nvidia K80/T4 GPU took around 8 hours where the majority of the time was taken by the process of question-answer pair creation. In the end of this fine-tuning cycle, we obtained a custom trained sentence transformer model to create numeric embeddings (dimension 768) of the text stored in our VDB.

**D2 Hyper-parameters for Generative LMs**

Note: temperature helps in controlling the creativity of a LM, and beam search pays more attention to word sequence when creating sentences than a default greedy search strategy leading to better answers in our context.

Table D1: Hyper-parameters used in Generative Answering

| **Hyper-parameter** | **Values** |
|---|---|
| min answer length | 40 |
| max answer length | 150 |
| Temperature | 0.7 |
| number of beams | 4 |

**D3 Semantic Text Similarity (STS)**

Table D2: STS score with respect to Gold-answer (overall)

| | | **Baseline-1** (ChatGPT) | **Baseline-2** [25] | **Baseline-3** [22] | **ELI5-ans** (joint-reasoning; model-1) | **LFQA-ans** (joint-reasoning; model-2) |
|---|---|---|---|---|---|---|
| **Gold Answer STS score (0-5)** | min | 0.40 | -0.04 | 0.15 | 0.21 | **1.86** |
| | max | 3.76 | 4.6 | 3.51 | 3.60 | **4.84** |
| | median | 2.66 | 2.12 | 2.48 | 2.31 | **3.04** |
| | mean | 2.53 (7.4e-06)[+] | 1.9 (8.0e-12)[+] | 2.33 (2.9e-10)[+] | 2.26 | **3.03** |
| | std. dev. | 0.71 | 1.2 | 0.62 | 0.78 | 0.60 |

***model-1***: answers obtained by our system using ELI5 model with joint-reasoning; ***model-2***: answers obtained by our system using LFQA model with joint-reasoning – our final model
[+] Values in brackets depict p-value of t-test conducted for all baselines with respect to model-2 (our final model)

Table D3: STS score with respect to Gold-answer (questions split by categories)

| | **Category** | **Baseline-1** (ChatGPT) | **Baseline-2** [25] | **Baseline-3** [22] | **ELI5-ans** (joint-reasoning; model-1) | **LFQA-ans** (joint-reasoning; model-2) |
|---|---|---|---|---|---|---|
| **Gold Answer STS median score (0-5)** | *about disease* | 3.10 | 2.96 | 2.97 | 2.97 | **3.38** |
| | *at risk* | 2.71 | 1.95 | 2.49 | 2.43 | **3.05** |
| | *cause* | 3.04 | 1.61 | 2.67 | 2.70 | **3.05** |
| | *diagnosis and test* | 3.15 | 1.93 | 2.06 | 2.31 | **3.15** |
| | *other* | 3.00 | 2.34 | 1.67 | 2.06 | **3.00** |
| | *symptom* | 3.05 | 2.73 | 2.66 | 2.08 | **3.05** |
| | *treatment* | 2.50 | 0.25 | 2.16 | 1.82 | **2.60** |

## E SENTENCE ENTAILMENT APPROACH

Specifically in our work, we performed a pair-wise entailment computation of each answer sentence in the system generated text with respect to each sentence in the gold answer text. We also did similar computation for answers from the ChatGPT system. A strict threshold of 0.95 was set to consider two sentences to be negatively entailed.

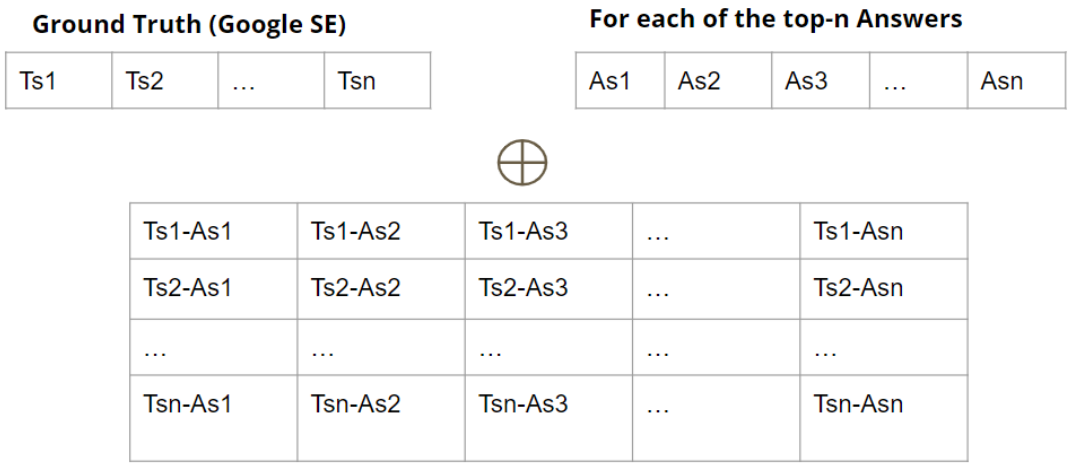

Figure D1: Pair-wise sentence entailment scoring between gold answer and machine generated answer

Tsn: nth sentence of the ground truth, Asn: nth sentence of the machine generated answer. A pairwise set is prepared and checked for any negative entailment between ground truth and system generated answer

## F ADDITIONAL EXPERIMENTS

### F1 Evaluation of KG responses for QA

We evaluated our KG (section 3.3.2) as a stand-alone QA resource to examine its ability to answer natural language questions. For this purpose, first we obtained semantically aware low-dimensional vector representations of the graph triples using the graph specific embedding techniques that facilitate downstream graph analytics and knowledge discovery tasks [16]. Specifically, in our work, we used a 2D embedding technique called ConvE [19].

While there are a number of embedding methods, in our work we used a common 1D (dimension) method, TransE [9] and a common 2D method, ConvE [19] for comparison. TransE is a 1D distance-based embedding technique, whereas ConvE uses a convolutional neural network to capture non-linear relations and generate embeddings based on 2D graph convolutions. The training involved the following split of dataset: for training (85%), for validation (5%), and for testing (10% triplets). A learning rate of 0.001, Adam optimizer, batch size of 10, max epochs of 1000 and negative log likelihood loss (NLL, a loss function for multi-class classification problems) were used for training. An early stopping criterion was used to stop training when no improvement was observed in the mean reciprocal rank (MRR) on the validation set.

For measuring the accuracy of graph embeddings in predicting subject or object entities, when one of them is missing, we used the popular metrics - Hits@N and MRR [49]. Hits@N ranks triplets and computes the percentage of positive triplets ranked in the top N results. MRR measures the average of reciprocal ranks of all the triplets. Higher values of these metrics indicate better accuracy of graph embeddings. Accordingly, we chose to use ConvE, as it performed better on both metrics (Table F1).

Table F1: Test Set Performance of Learned Embeddings for KG

| Method | Hits@1 | Hits@10 | Hits@100 | MRR |
|--------|--------|---------|----------|------|
| *TransE* | 0.01 | 0.05 | 0.24 | 0.02 |
| *ConvE* | **0.17** | **0.17** | **0.22** | **0.18** |

A KG has a well-defined structure (h, r, t) and KG embeddings further capture the semantic relationships between nodes and edges facilitating the triplet completion task. We used this fact for querying our KG. *First*, from a given natural language question we extracted the structure, using a custom program we wrote in Python, to see if it is of the form **< h, r, ? >** OR **< ?, r, t >** OR **< h, ?, t >**. We then used this structure to translate a question to a triplet format to query our KG. For assessing answer accuracy, we tested our KG with factual questions. An answer would score 1 (if correct), 0.5 (if partially correct), or 0 otherwise. While the KG showed good accuracy of 75% for factual QA, it performed poorly on long questions (e.g., what are the reasons for eczema?) with 25% accuracy. This led us to not use the KG as standalone QA resource but rather use it as a resource to refine answers using the fact that, the KG has trustworthy factual information in its nodes from Mayo Clinic data.

**F2 Evaluating Quality of Fine-Tuned Sentence Transformer Model**

It is important to evaluate the encoding quality of our newly domain adapted sentence transformer model (section 3.3.3) before we can use it in our proposed system. To assess the quality of the encodings produced, we compared it with the top-3 pre-trained sentence transformers available with necessary details on the sentence transformer website [58]. In order of their ranking, these models are namely – (1) *all-mpnet-base-v2,* (2) *multi-qa-distilbert-cos-v1*, and (3) *multi-qa-MiniLM-L6-cos-v1*. Each of the models is specifically used for semantic parsing which is also our system's requirement and are globally ranked as top improvements achieved over their respective base models. Hence they serve as good references for comparison. Since this fine-tuning of all the top-3 sentence transformers largely involved non-medical text, we decided to fine-tune the Microsoft/mpnet-base model on our own collected and cleaned lay user friendly data.

Table F2: Comparison of our Sentence Transformer model with other pre-trained models (STS Benchmark Dataset)

| Sentence Transformer* | all-mpnet-base-v2 | multi-qa-distilbert-cos-v1 | multi-qa-MiniLM-L6-cos-v1 | *our sentence transformer* |
|---|---|---|---|---|

| Base-model | microsoft/mpnet-base | distilbert-base | nreimers/MiniLM-L6-H384-uncased | microsoft/mpnet-base |
|---|---|---|---|---|
| Training Data | 1B+ training pairs | 215M QA pairs | 215M QA pairs | 31, 628 QA pairs |
| Count of data Sources | 32 | 15 | 15 | 2 |
| Correlation with STS label | 0.88 | 0.79 | 0.8 | **0.83** |
| Data Sources details | https://huggingface.co/sentence-transformers/all-mpnet-base-v2 | https://huggingface.co/sentence-transformers/multi-qa-distilbert-cos-v1 | https://huggingface.co/sentence-transformers/multi-qa-MiniLM-L6-cos-v1 | Mayo Clinic + NIH |

The comparison of our model performance, with respect to the other top-3 models, was done using the publicly available Semantic Text Similarity (STS) benchmarking dataset [13]. It is a collection of English datasets used in all the STS tasks organized in SemEval[9] and it contains text from news, captions, and user forums. Particularly, this dataset contains a set of 1500 sentence pairs with features such as ('sentence1', 'sentence2', 'label', 'idx') where *'label'* is the numeric score for the extent of similarity (between sentences1/2) on a scale of 0-5. The similarity score for each sentence pair is carefully assigned by human evaluators and hence it is widely used for development and testing of NLP systems.

We computed the correlation of similarity scores of each sentence pair given in the dataset and the score obtained from various sentence transformer models including our own custom trained sentence transformer (see Table 3). Our fine-tuned sentence transformer achieved a good similarity score of 0.83 in producing sentence encodings which is comparable to the other three sentence transformer models fine-tuned on non-medical texts. Hence, we decided to use our own sentence transformer model for two reasons - (1) it produces high quality sentence encodings for sematic searches which correlated strongly with the scores of human evaluators, and (2) it is domain adapted specifically on our own data encompassing the knowledge of medical text from the whole VDB which is essential when producing numerical encodings of natural questions presented to our proposed system.

**F3 Length-controlled Comparison of Answer Quality**

We did a length-controlled comparison of answer quality with the ChatGPT system. We observed that the average answer length of ChatGPT was around 150 words with some answers containing more than 250 words. Since metrics such as ROUGE, STS, BERTScore use the text contents (individual words, n-grams, or complete sentences) to compute their scores, longer answers produced by any system may have an unfair advantage when content overlap is being measured against a gold answer. Factoring this, we did a length-controlled comparison of answer quality by only considering answers less than or equal to 150 words long (i.e., filtered set on ChatGPT responses; total 59 questions). Our system achieved a mean ROUGE score of 0.43 (± 0.24) vs 0.22 (± .08) for ChatGPT, mean STS score of 3.13 (± 0.6) vs 2.69 (± 0.6) for ChatGPT, and mean BERTScore of 0.90 (± 04) vs 0.86 (± 0.02) for ChatGPT. All mean values were found to be statistically significant ($p < 0.05$) when compared to respective ChatGPT scores using a t-test. Detailed scores of this additional experiment are provided below in the Tables F3.1- F3.6.

---

[9] https://semeval.github.io/

Table F3.1: Rouge-L (F1-score) with respect to Gold-answer (overall)

|  |  | **Baseline-1** (ChatGPT) | **ELI5-ans** (joint-reasoning; model-1) | **LFQA-ans** (joint-reasoning; model-2) |
|---|---|---|---|---|
| **Gold Answer ROUGE-L F1-score (0-1)** | min | 0.00 | 0.04 | **0.08** |
|  | max | 0.48 | 0.79 | **0.96** |
|  | median | 0.22 | 0.19 | **0.38** |
|  | mean | 0.22 (2.7e-08)[+] | 0.26 (4.9e-05)[+] | **0.43** |
|  | std. dev. | 0.08 | 0.19 | 0.24 |

*model-1*: answers obtained by our system using ELI5 model with joint-reasoning; *model-2*: answers obtained by our system using LFQA model with joint-reasoning
+ Values in brackets depict p-value of t-test conducted for all baselines with respect to model-2 (our final model)

Table F3.2: Rouge-L (F1-score) with respect to Gold-answer (questions split by categories)

|  | **Category** | **Baseline-1** (ChatGPT) | **ELI5-ans** (joint-reasoning; model-1) | **LFQA-ans** (joint-reasoning; model-2) |
|---|---|---|---|---|
| **Gold Answer median ROUGE-L F1-score (0-1)** | *about disease* | 0.27 | 0.29 | **0.42** |
|  | *at risk* | 0.23 | 0.17 | **0.40** |
|  | *Cause* | 0.25 | 0.24 | **0.29** |
|  | *diagnosis and test* | 0.2 | 0.39 | **0.72** |
|  | *Other* | 0.17 | 0.10 | **0.46** |
|  | *symptom* | 0.19 | 0.17 | **0.32** |
|  | *treatment* | 0.2 | 0.10 | **0.31** |

Table F3.3: STS score with respect to Gold-answer (overall)

|  |  | **Baseline-1** (ChatGPT) | **ELI5-ans** (joint-reasoning; model-1) | **LFQA-ans** (joint-reasoning; model-2) |
|---|---|---|---|---|
| **Gold Answer STS score (0-5)** | Min | 0.43 | 0.69 | **1.85** |
|  | Max | 3.76 | 3.59 | **4.84** |
|  | Median | 2.85 | 2.52 | **3.10** |
|  | Mean | 2.69 (0.00)[+] | 2.45 (3.1e-07)[+] | **3.13** |

|  | | std. dev. | 0.68 | 0.69 | 0.68 |

+ *Values in brackets depict p-value of t-test conducted for all baselines with respect to model-2 (our final model)*

Table F3.4: STS score with respect to Gold-answer (questions split by categories)

|  | Category | Baseline-1 (ChatGPT) | ELI5-ans (joint-reasoning; model-1) | LFQA-ans (joint-reasoning; model-2) |
| --- | --- | --- | --- | --- |
| **Gold Answer STS median score (0-5)** | *about disease* | 3.10 | 2.97 | **3.38** |
|  | *at risk* | 2.79 | 2.35 | **2.98** |
|  | *cause* | 2.93 | 2.76 | **3.04** |
|  | *diagnosis and test* | 2.86 | 2.77 | **3.63** |
|  | *other* | 2.47 | 2.17 | **3.08** |
|  | *symptom* | 2.52 | 2.3 | **3.12** |
|  | *treatment* | 2.57 | 1.93 | **3.02** |

Table F3.5: BERTScore with respect to Gold-answer (overall)

|  |  | Baseline-1 (ChatGPT) | ELI5-ans (joint-reasoning; model-1) | LFQA-ans (joint-reasoning; model-2) |
| --- | --- | --- | --- | --- |
| **Gold Answer BERTScore (0-1)** | min | 0.80 | 0.82 | **0.84** |
|  | max | 0.90 | 0.95 | **0.99** |
|  | median | 0.86 | 0.87 | **0.90** |
|  | mean | 0.86 (1.4e-10)+ | 0.87 (3.5e-05)+ | **0.90** |
|  | std. dev. | 0.02 | 0.03 | 0.04 |

+ *Values in brackets depict p-value of t-test conducted for all baselines with respect to model-2 (our final model)*

Table F3.6: BERTScore with respect to Gold-answer (questions split by categories)

|  | Category | Baseline-1 (ChatGPT) | ELI5-ans (joint-reasoning; model-1) | LFQA-ans (joint-reasoning; model-2) |
| --- | --- | --- | --- | --- |
| **Gold Answer median BERTScore (0-1)** | *about disease* | 0.88 | 0.90 | **0.93** |
|  | *at risk* | 0.86 | 0.86 | **0.89** |
|  | *cause* | 0.85 | 0.88 | **0.88** |
|  | *diagnosis and test* | 0.86 | 0.89 | **0.94** |
|  | *other* | 0.86 | 0.87 | **0.91** |
|  | *symptom* | 0.85 | 0.85 | **0.88** |
|  | *treatment* | 0.86 | 0.84 | **0.89** |

**F4 Generating Answers without a VDB (Ablation)**

Our VDB (see step-1; Figure 1) is an important part of the proposed system. To highlight its importance, we generated answers to all the questions without using data from VDB to provide an explicit context for answer generation during the answer formation stage. It can be seen from Table F4 that without accompanying good quality contexts from a VDB, the large LMs run the risk of providing confusing, or incorrect answers. This underpins the importance of using a good quality VDB capable of providing relevant contextual information for answer generation to a large LM.

Table F4: Generative Answer Quality without a Supporting VDB to Provide Relevant Context

| Question | No VDB answer | Model-2 (joint-reasoning) |
| --- | --- | --- |
| How to diagnose Chronic Cough? | Coughing can be **caused by** a number of things. The most common cause of cough is a cough reflex. Cough reflexes are: 1. Your body produces mucus. 2. You cough. 3. Blood vessels constrict. 4. There are other symptoms. | There are few ways to **diagnose** Chronic Cough: Blood tests, sputum mucus test, imaging tests such as chest x-ray or CT scan, spirometry or methacholine challenge test. |
| What are symptoms of Heart Failure? | Heart failure is not a disease. It is a medical condition. The symptoms of heart failure are: 1. Cardiovascular failure. 2. Heart attack. 3. Stroke. 4. Congestion. 5. Fatigue. 6. Seizures. | The symptoms of heart failure are: Chest pain, shortness of breath, fatigue and weakness. Rapid or irregular heartbeat. Swelling in legs, ankles and feet. Persistent cough or wheezing with white or pink blood-tinged mucus. Reduced ability to exercise. |
| How to diagnose Osteoarthritis? | Osteoarthritis is not a disease. It is a condition that is caused by a degeneration of cartilage. There are two types of osteoblasts: osteoclasts and osteoporosis. The first type is called osteoblastoma. | There are a lot of different types of osteoarthritis, and there are many different ways to diagnose it. One way is to look at X-rays. X-rays, which can show loss of joint space, bone remodeling and bone spurs. Magnetic resonance imaging (MRI) is another way. |